\documentclass{article}
\pdfoutput=1 
\usepackage[final, nonatbib]{nips_2016} %
\usepackage[utf8]{inputenc} 
\usepackage[T1]{fontenc}    
\usepackage{hyperref}       
\usepackage{url}            
\usepackage{booktabs}       
\usepackage{amsfonts}       
\usepackage{nicefrac}       
\usepackage{microtype}      

\usepackage{graphicx}

\title{Network Trimming: A Data-Driven Neuron Pruning Approach towards Efficient Deep Architectures}

\author{
  Hengyuan Hu \thanks{Part of the work was done when Hengyuan Hu and Rui Peng were interns in SenseTime Group Limited} \\
  HKUST \\
  \texttt{hhuaa@ust.hk} \\
  \And
  Rui Peng \footnotemark[1] \\
  HKUST \\
  \texttt{rpeng@ust.hk} \\
  \And
  Yu-Wing Tai \\
  SenseTime Group Limited \\
  \texttt{yuwing@sensetime.com} \\
  \And
  Chi-Keung Tang \\
  HKUST \\
  \texttt{cktang@cse.ust.hk} \\
}

\begin{document}

\maketitle

\begin{abstract}
State-of-the-art neural networks are getting deeper and wider. While their performance increases with the increasing number of layers and neurons, it is crucial to design an efficient deep architecture in order to reduce computational and memory costs. Designing an efficient neural network, however, is labor intensive requiring many experiments, and fine-tunings. In this paper, we introduce network trimming which iteratively optimizes the network by pruning unimportant neurons based on analysis of their outputs on a large dataset. Our algorithm is inspired by an observation that the outputs of a significant portion of neurons in a large network are mostly zero, regardless of what inputs the network received. These zero activation neurons are redundant, and can be removed without affecting the overall accuracy of the network. After pruning the zero activation neurons, we retrain the network using the weights before pruning as initialization. We alternate the pruning and retraining to further reduce zero activations in a network. Our experiments on the LeNet and VGG-16 show that we can achieve high compression ratio of parameters without losing or even achieving higher accuracy than the original network.
\end{abstract}

\section{Introduction}
Neural networks have been widely adopted in many scenarios, achieving state-of-the-art results in numerous tasks~\cite{alex}~\cite{speech-recog}. One of the keys to improved performance is their increased depth and width and thus the increased number of parameters. In computer vision, we have witnessed orders of magnitude increase in the number of parameters in CNNs from LeNet with less than 1M parameters in handwritten digit classification~\cite{lenet} to Deepface with more than 120M parameters in human face classification~\cite{deepface}.

Although CNNs with elegant network architectures are easy to deploy in real-world tasks, designing one can be hard and labor-intensive, which involves significant amount of effort in empirical experiments. In terms of designing the network architecture, one crucial part is to determine the number of neurons in each layer. There is no way to directly arrive at an optimal number of neurons for each layer and thus even the most successful network architectures use empirical numbers like 128, 512, 4096. Experienced scientists often arrive at the numbers once they deem the network have enough representation power for the specific task. However, the extremely sparse matrices produced by top layers of neural networks have caught our attention, indicating that empirically designed networks are heavily oversized. After some simple statistics, we find that many neurons in a CNN have very low activiations no matter what data is presented. Such weak neurons are highly likely to be redundant and can be excluded without damaging the overall performance. Their existence can only increase the chance of overfitting and optimization difficulty, both of which are harmful to the network.

With the motivation of achieving more efficient network architectures by finding the optimal number of neurons in each layer, we come up with an iterative optimization method that gradually eliminates weak neurons in a network via a pruning-retraining loop. Starting from an empirically designed network, our algorithm first identifies redundant weak neurons by analyzing their activiations on a large validation dataset. Then those weak neurons are pruned while others are kept to initialize a new model. Finally, the new model is retrained or fine-tuned depending on the performance drop. The retrained new model can maintain the same or achieve higher performance with smaller number of neurons. This process can be carried out iteratively until a satisfying model is produced.


\section{Related Work}
Significant redundancy has been demonstrated in several deep learning models~\cite{pred_param} and such redundancy is mainly caused by the overwhelming amount of parameters in deep neural networks. An over-parameterized model not only wastes memory and computation, but also leads to serious overfitting problem. Therefore, reducing the number of parameters has been studied by many researchers in this field. However, there is little work directly addressing the optimization of the number of neurons. Most previous works on improving network architectures fall in two main categories; one concentrates on the high level architectural design and the other focuses on low level weight pruning.

On the high level side, some researchers invented new layers or modules to substitute main bottleneck components in conventional neural networks. Two famous examples of this kind are the global average pooling in Network in Network~\cite{network_in_network} invented to replace the extremely dense paramaterized fully connected layer and the inception module employed by GoogLeNet~\cite{googlenet} to avoid explosion in computational complexity at later stage. Both methods achieve state-of-the-art results on several benchmarks with much less memory and computation consumption. More recently, SqueezeNet~\cite{squeeze-net} used a Fire module together with other strategies to achieve AlexNet-level accuracy with $50\times$ less parameters.

On the low level side, different methods have been explored to reduce number of connections and weights in neural networks. Some late 20th century methods, such as magnitude-based approach~\cite{mag_based_pruning} and Hessian matrix based approach~\cite{optimal_brain_surgeon}, prune weights basing on numerical properties of the weights and loss functions without any external data involved. Han et al. recently proposed an iterative method~\cite{learning_connections} to prune connections in deep architectures, together with an external compression by quantization and encoding~\cite{deep_compress}. The network is first pruned by removing low weights connections. Then, learned mapping of similar weights to fixed bits are used to perform quantization of weights after pruning, which facilitates the Huffman coding compression in the last stage to reduce bits for storage. When all three techniques used in pipeline, the number of parameters in the network can be reduced by around $10\times$.

While above methods work well in practice by reducing number of parameters directly, we seek answers to the fundamental problem that lies in the middle of those two levels of approaches -- determining the optimal number of neurons for each layer for a given network architecture and specific tasks. Along our direction, not only can we achieve parameter savings without the need of seeking new network architectures, we can also evaluate the redundancy in each layer of a network, and thus provide guidance on effective ways for architecture optimization in large neural networks.

\section{Zero Activations and Network Trimming}
In this section, we describe our algorithm for network trimming. To facilitate our discussions, we use VGG-16~\cite{vgg} as our case study. The VGG-16 network consists of 13 convolutional layers, and 3 fully connected layers. Each of the layers is followed by a ReLU~\cite{relu} layer for non-linear mapping. The VGG-16 is recognized as one of the representative network which has been adopted to many applications~\cite{faster}~\cite{video-to-text}, not limited to object classification and localization tasks.

\subsection{Zero Activations in VGG-16}
\label{analysis}
We define Average Percentage of Zeros (APoZ) to measure the percentage of zero activations of a neuron after the ReLU mapping. Let 
$O^{(i)}_{c}$ denotes the output of $c$-th channel in $i$-th layer, our $APoZ^{(i)}_{c}$ of the $c$-th neuron in $i$-th layer is defined as:
\begin{equation}\label{eq:APoZ}
APoZ^{(i)}_{c} = APoZ(O^{(i)}_{c})= \frac{\sum^{N}_k\sum^{M}_jf(O^{(i)}_{c,j}(k) = 0)}{N\times M}
\end{equation}
where $f(\cdot)=1$ if true, and $f(\cdot)=0$ if false, $M$ denotes the dimension of output feature map of $O^{(i)}_{c}$, and $N$ denotes the total number of validation examples. The larger number of validation examples, the more accurate is the measurement of APoZ. In our experiment, we use the validation set ($N=50,000$) of ImageNet classification task to measure APoZ.

We use the definition of APoZ to evaluate the importance of each neuron in a network. To validate our observation that the outputs of some neurons in a large network are mostly zero, we calculate the APoZ of each neuron and find that there are $631$ neurons in the VGG-16 network which have APoZ larger than $90\%$. 

\begin{table}[h]
  \caption{Mean APoZ of each layer in VGG-16}
  \label{vgg_apoz}
  \centering
  \begin{tabular}{llllll}
    \toprule
Layer          & CONV1-1 & CONV1-2 & CONV2-1 & CONV2-2 & CONV3-1 \\
Mean APoZ (\%) & 47.07   & 31.34   & 33.91   & 51.98   & 47.93   \\
\midrule
Layer          & CONV3-2 & CONV3-3 & CONV4-1 & CONV4-2 & CONV4-3 \\
Mean APoZ (\%) & 48.84   & 69.93   & 65.33   & 70.76   & 87.30   \\
\midrule
Layer          & CONV5-1 & CONV5-2 & CONV5-3 & FC6     & FC7     \\
Mean APoZ (\%) & 76.51   & 79.73   & 93.19   & 75.26   & 74.14   \\
    \bottomrule
  \end{tabular}
\end{table}

\begin{figure}[h]
\minipage{0.45\textwidth}
  \includegraphics[width=\linewidth]{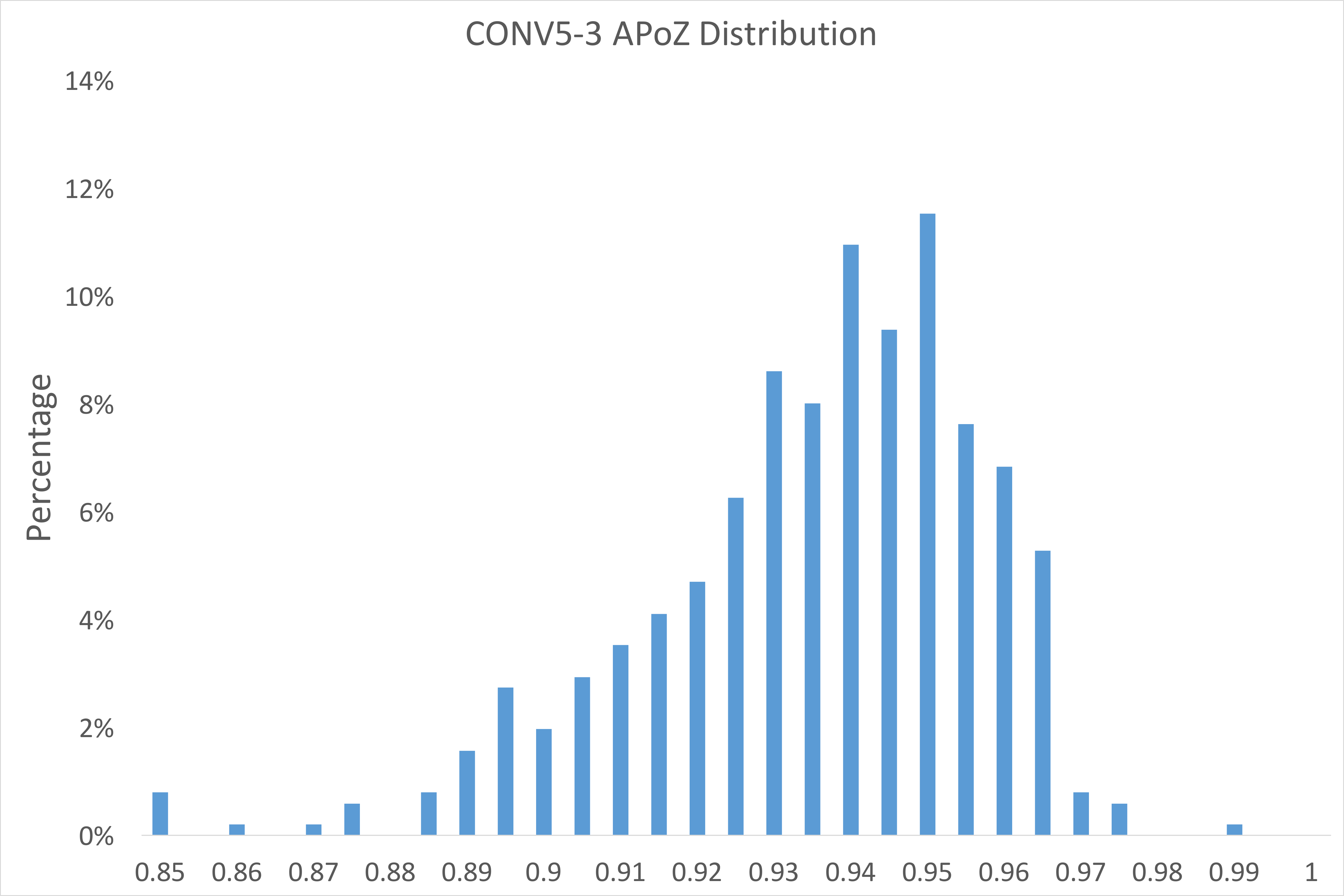}
  \caption{CONV5-3 APoZ Distribution}\label{fig:conv53-apoz}
\endminipage\hfill
\minipage{0.45\textwidth}
  \includegraphics[width=\linewidth]{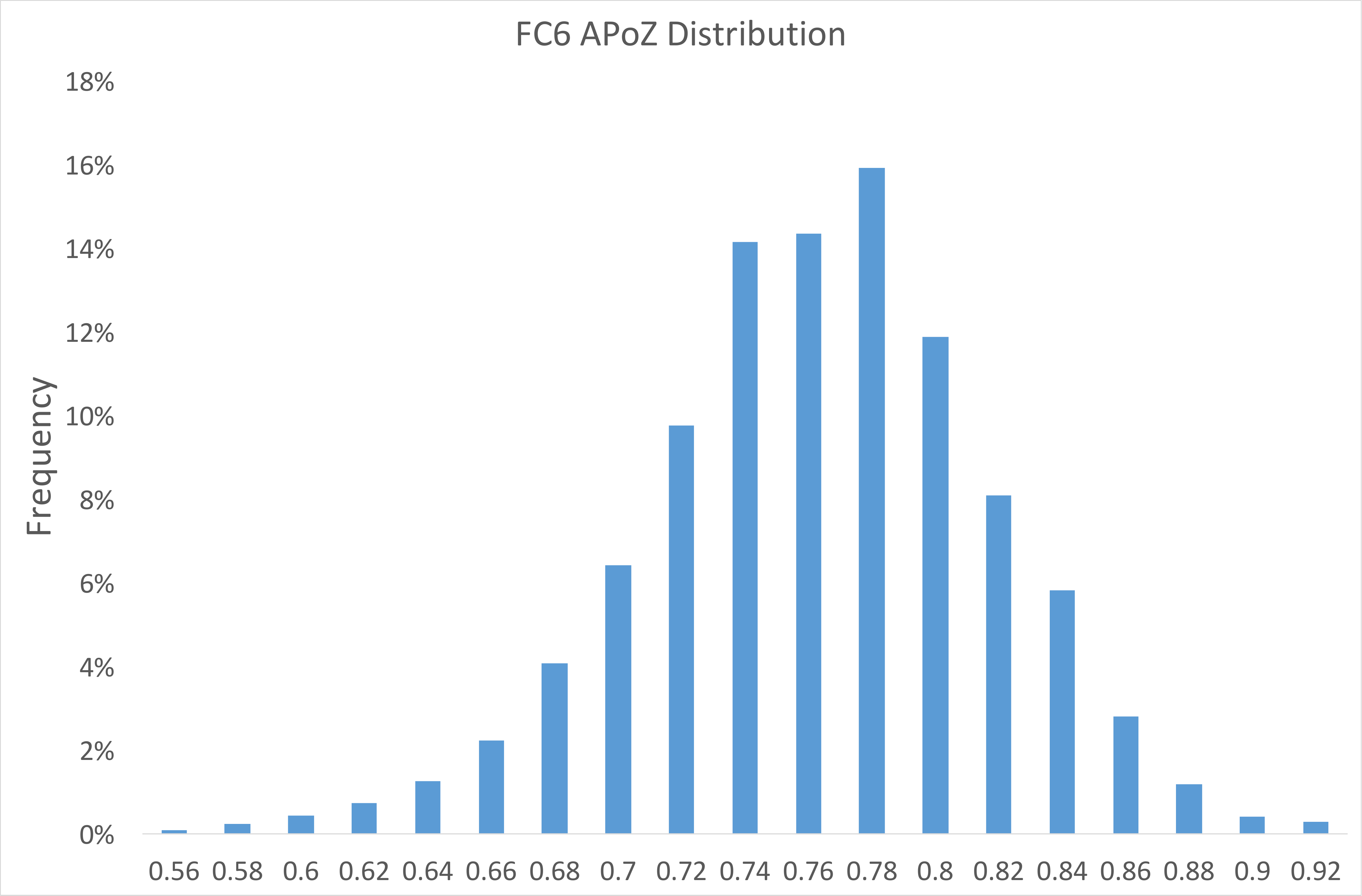}
  \caption{FC6 APoZ Distribution}\label{fig:fc6-apoz}
\endminipage
\end{figure}

To better understand the behavior of zero activations in a network, we compute the mean APoZ (Table~\ref{vgg_apoz}) of all neurons in each layer (except for the last one) of the VGG-16 network. Since the VGG-16 network has inverse pyramid shape, most redundancy occurs at the higher convolutional layers and the fully connected layers. The higher mean APoZ also indicates more redundancy in a layer. Detailed distributions of APoZ of 512 CONV5-3 neurons and 4096 FC6 neurons are shown in Figure~\ref{fig:conv53-apoz},~\ref{fig:fc6-apoz} respectively. Since a neural network has a multiplication-addition-activation computation process, a neuron which has its outputs mostly zeros will have very little contribution to the output of subsequent layers, as well as to the final results. Thus, we can remove those neurons without harming too much to the overall accuracy of the network. In this way, we can find the optimal number of neurons for each layer and thus obtain a better network without redesign and extensive human labor.

\subsection{Network Trimming and Retraining}
\label{core}
\begin{figure}[h]
\minipage{0.32\textwidth}
  \includegraphics[width=\linewidth]{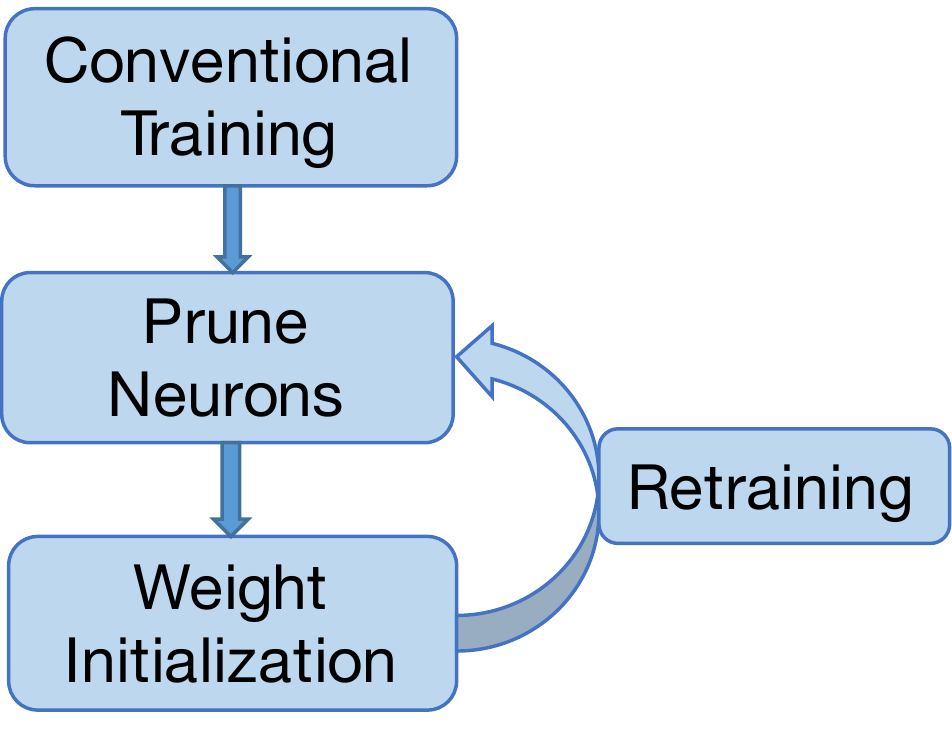}
  \caption{Three main steps for trimming}\label{fig:process}
\endminipage\hfill
\minipage{0.32\textwidth}
  \includegraphics[width=\linewidth]{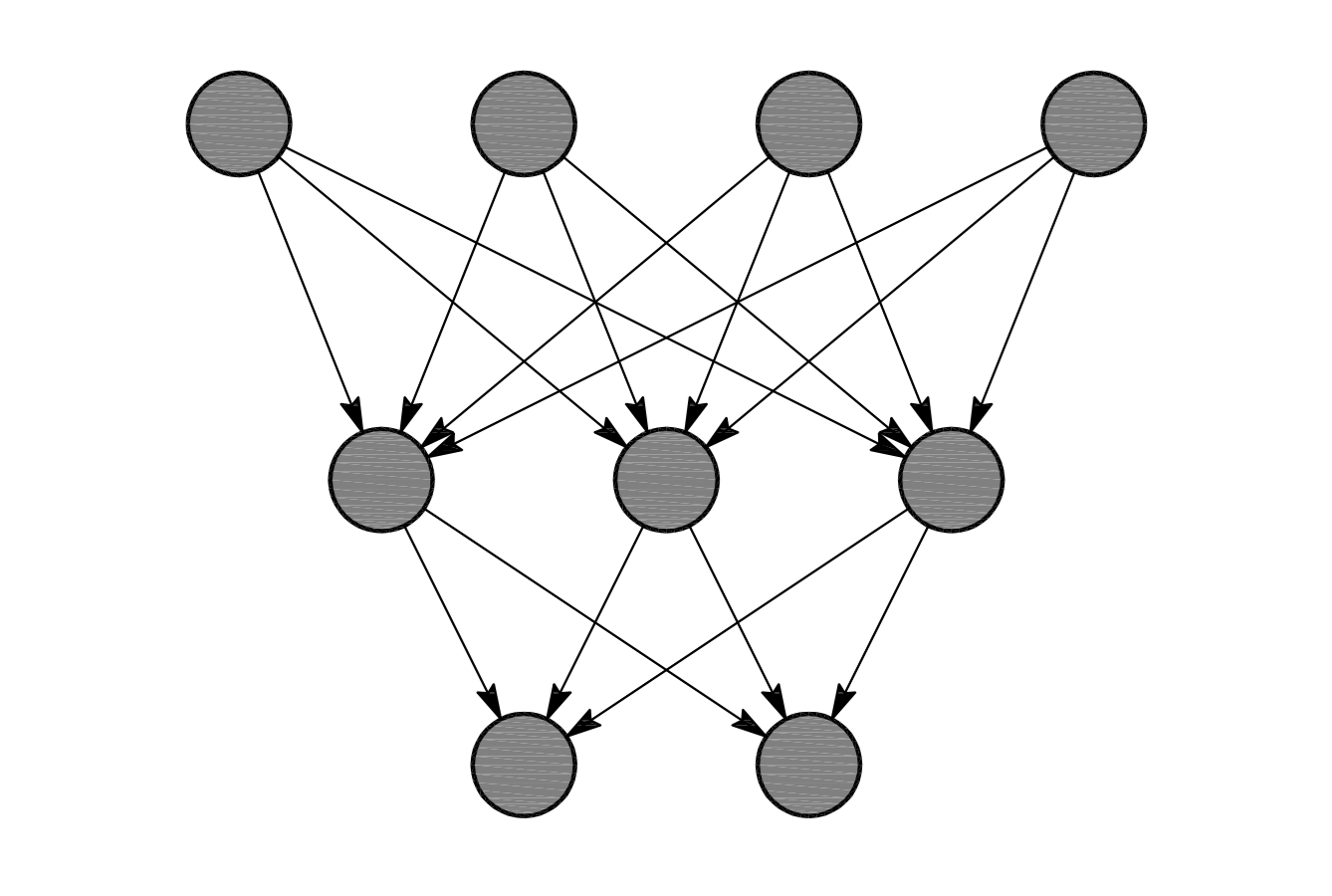}
  \caption{Before pruning}\label{fig:prune-before}
\endminipage
\minipage{0.32\textwidth}
  \includegraphics[width=\linewidth]{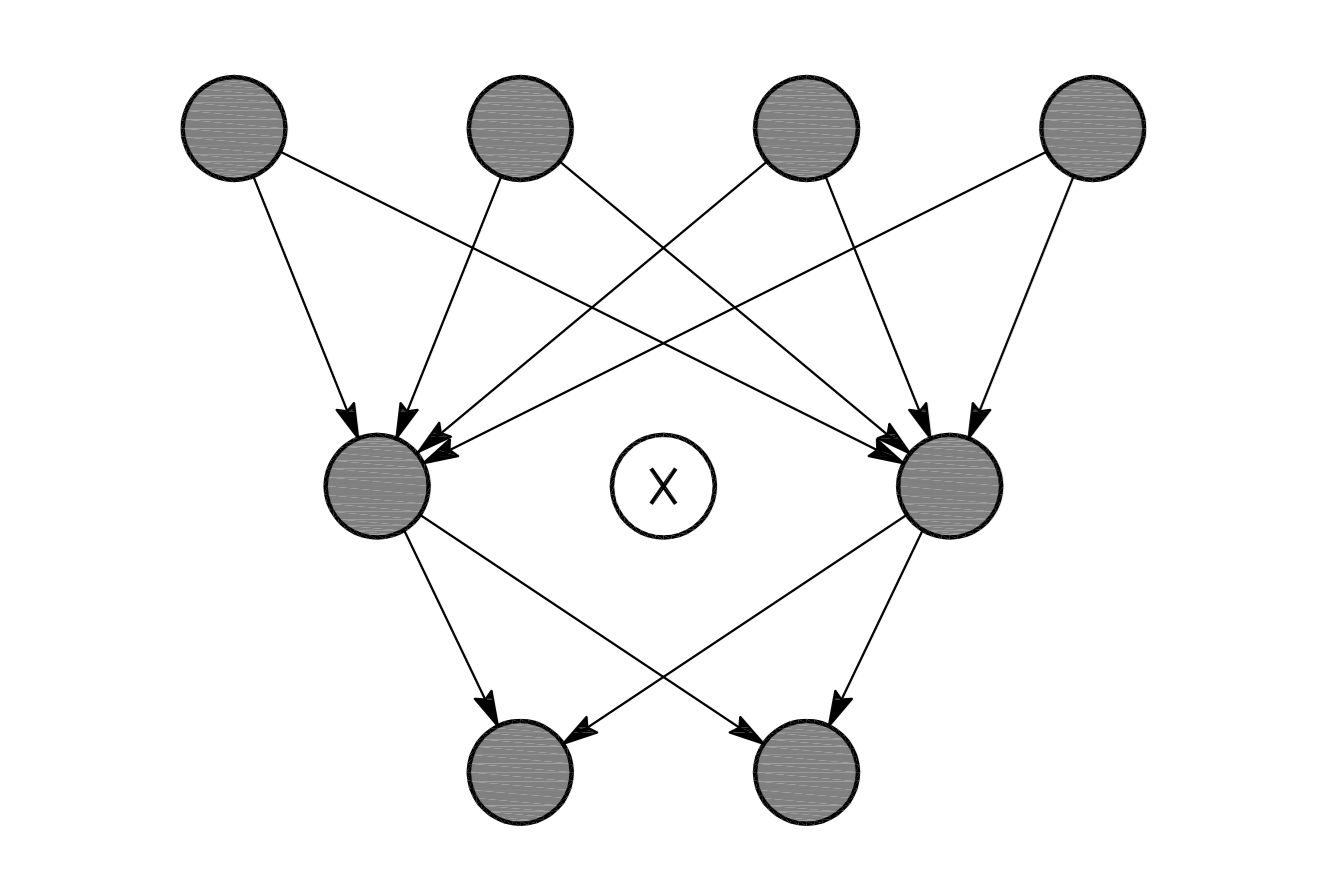}
  \caption{After pruning}\label{fig:prune-after}
\endminipage
\end{figure}

Our network trimming method consists of three main steps, as illustrated in Figure~\ref{fig:process}. First the network is trained under conventional process and the number of neurons in each layer is set empirically. Next, we run the network on a large validation dataset to obtain the APoZ of each neuron. 

Neurons with high APoZ are pruned according to certain criteria. The connections to and from the neuron are removed accordingly when a neuron is pruned (see Figure~\ref{fig:prune-before}~\ref{fig:prune-after}). After the neuron pruning, the trimmed network is initialized using the weights before trimming. The trimmed network exhibits some level of performance drop. Thus, in the final step, we retrain the network to strengthen the remaining neurons to enhance the performance of the trimmed network. 



The weight initialization is necessary for the network to obtain the same performance as it was before the trimming. If a trimmed network is trained from scratch, we find that it contains larger percentage of zero activation neurons than the counterpart with weight initialization. This means that a retrained network without weight initialization is much less efficient.

We experimented different ways to prune the neurons according to the APoZ measurements. We found that pruning too many neurons at once severely damaged the performance, and the performance drops are unrecoverable. Therefore, we chose an iterative scheme to trim a network. However, it is not trivial to trim a network with deep architecture. If too many layers are trimmed in one step, the performance would drop by a large margin, and it is hard to recover the original performance before trimming through the retraining. For example, trimming CONV4, CONV5, FC6 and FC7 of the VGG-16 network concurrently would lead to a $46.650\%$ top-5 accuracy in the image classification task, where the original accuracy of VGG-16 \footnote{Single scale, without dense evaluation~\cite{vgg}} is $88.444\%$. On the other hand, if only the CONV5-3 and FC6 are trimmed, the trimmed network with weight initialization before retraining can achieve $85.900\%$ top-5 accuracy. After retraining, the trimmed network achieves $90.278\%$ accuracy which is even higher than the original accuracy before trimming.

Empirically, we found that starting to trim from a few layers with high mean APoZ, and then progressively trim its neighboring layers can rapidly reduce the number of neurons while maintaining the performance of the original network. To decide which neurons to prune, we empirically found that pruning the neurons whose APoZ is larger than one standard derivation from the mean APoZ of the target trimming layer would produce good retraining results. Using this threshold, we would reject $16\%$ of neurons on average from the trimmed layers, assuming that the APoZ values roughly follow a Gaussian distribution.

\section{Experiments}
We implemented our algorithm using the standard Caffe~\cite{caffe} library. To obtain the weights for initialization for retraining, we use the Python and PyCaffe interface to copy the weights of remaining connections after the trimming. We tested our algorithm primarily on two networks, LeNet~\cite{lenet} on MNIST dataset and VGG-16 on ImageNet classification dataset~\cite{imagenet}.

\subsection{LeNet}
The LeNet network consists of two convolutional layers followed by two fully connected layers, the layers have $20,50,500,10$ outputs respectively. We use a short hand notion (20-50-500-10) to denote the number of neurons in each layer of the network. In the LeNet, $93\%$ of parameters are in the connections between the CONV2 layer and the FC1 layer. Consequently, we can easily achieve a more efficient network by trimming the size of CONV2 and FC1 layers.

\subsubsection{Effectiveness}

We apply our algorithm to iteratively prune the neurons in CONV2 and FC1 layers, as shown in Table~\ref{lenet_performance_with_init}. At the first iteration of the pruning, the numbers of neurons in CONV2 and FC1 layers are reduced to 41 and 426 respectively, which achieves $1.41\times$ compression on the number of parameters after the first pruning. The accuracy drops from $99.27\%$ to $98.75\%$ after the pruning, but before retraining. After retraining the network, we achieve $99.29\%$ accuracy which is slightly higher than the original accuracy. We repeat these processes for 4 iterations. As shown in Table~\ref{lenet_performance_with_init}, our algorithm achieves more than 2$\sim$ 3$\times$ compression on the number of parameters without loss in accuracy.

\begin{table}[h]
  \caption{Iterative Trimming on LeNet}
  \label{lenet_performance_with_init}
  \centering
  \begin{tabular}{llll}
    \toprule
    Network Config & Compression Rate & Initial Accuracy (\%) & Final Accuracy (\%)\\
    \midrule
    (20-50-500-10) & 1.00 & 10.52  & 99.31 \\
    (20-41-426-10) & 1.41 & 98.75  & 99.29 \\
    (20-31-349-10) & 2.24 & 95.34  & 99.30 \\
    (20-26-293-10) & 3.11 & 88.21  & 99.25 \\
    (20-24-252-10) & 3.85 & 96.75  & 99.26 \\
    \bottomrule
  \end{tabular}
\end{table}

\subsubsection{Necessity of Weight Initialization}
\label{lenet_weight_init}
We experiment our algorithm with retraining with and without weight initialization, as summarized in Table~\ref{lenet_performance_without_init}. The network exhibits deterioration in classification accuracy without weight initialization, whereas with proper weight initialization from the ancestor network from the previous iteration, the trimmed network can retain its original or even achieve higher accuracy.

\begin{table}[h]
  \caption{Iterative Trimming on LeNet with and without Weight Initialization}
  \label{lenet_performance_without_init}
  \centering
  \begin{tabular}{llll|lll}
    \toprule
    \multicolumn{1}{c}{}& \multicolumn{3}{c}{With Weight Init} & \multicolumn{3}{c}{Without Weight Init} \\
    \cmidrule{2-7}
    Number of Neurons in FC1 & 500 & 426 & 349 & 500 & 420 & 303 \\
    Accuracy (\%)      & 99.31 & 99.29 & 99.30 & 99.31 & 99.23 & 99.18 \\
    Mean APoZ (\%)     & 52.30 & 45.85 & 42.70 & 52.30 & 55.08 & 55.08 \\
    \#\{APoZ>0.6\}     & 154   & 77    & 17    & 154   & 160   & 110  \\
    \#\{APoZ>0.7\}     & 87    & 10    & 0     & 87    & 102   & 78   \\
    \#\{APoZ>0.8\}     & 54    & 0     & 0     & 54    & 61    & 49   \\
    \#\{APoZ>0.9\}     & 33    & 0     & 0     & 33    & 40    & 32   \\
    \bottomrule
  \end{tabular}
\end{table}

Moreover, we observe that with the weight initialization, the trimmed network consistently has smaller mean APoZ values than its ancestor network. This means that the retrained network has less redundancy than its ancestor network. In contrast, mean APoZ values increase if we retrain the network from scratch even though the trimmed network has less neurons than its ancestor network. This observation gives us an insight that proper weight initialization is necessary to achieve an efficient trimmed network.

\subsection{VGG-16}
\label{experiment_vgg}

\subsubsection{Effectiveness}
With the similar objective to obtain optimal number of neurons in each layer, we analyzed the APoZ values of $O^{(i)}_{c}$ for all $i$ and $c$ in VGG-16 on ImageNet classification validation set. As shown in Table~\ref{vgg_apoz}, CONV4, CONV5 and FC layers have higher mean APoZ compared with bottom layers, exhibiting more redundancy. Drawing from previous experience on LeNet, we focus on the parameter bottleneck of VGG-16. We trim the VGG-16 network starting from the CONV5-3 and FC6 layers since they account for 100M/138M  parameters. 

\begin{figure}[h]
  \centering
  \includegraphics[width=0.6\linewidth]{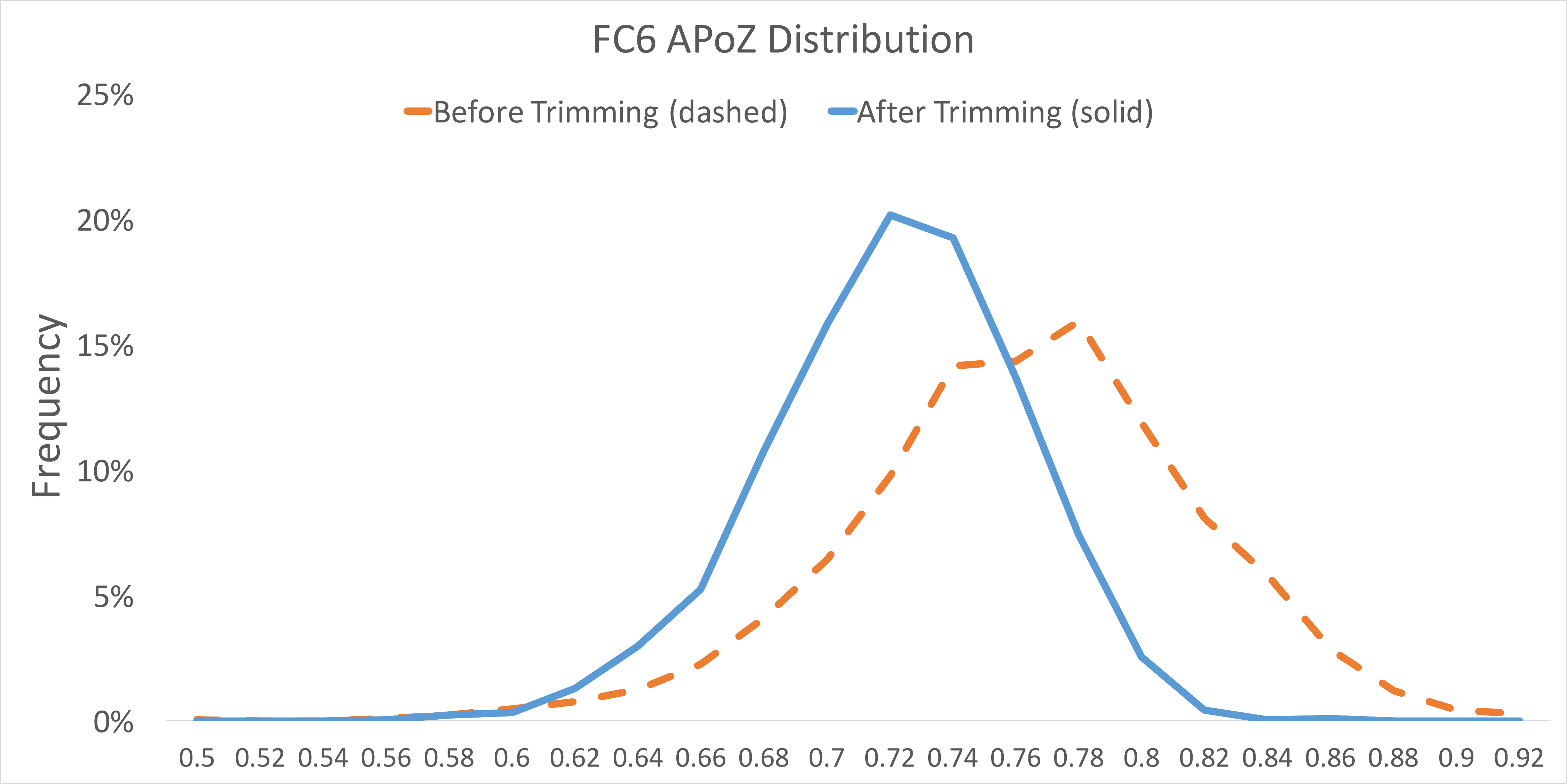}
  \caption{FC6 APoZ distribution before and after trimming}
  \label{fig:fc6-apoz-trim}
\end{figure}

We iteratively prune neurons from CONV5-3 and FC6 layers. Similar to the case in LeNet, the trimming process can effectively eliminate neurons with high APoZ. As shown in Figure~\ref{fig:fc6-apoz-trim}, after trimming, the entire distribution of APoZ in $O^{(fc6)}$ shifts left, indicating a significant drop in network redundancy. Meanwhile, the diminishing tail on the right side of the curve manifests that the weak neurons in FC6 are vanishing, a proof of the benefit gained from weight initialization as discussed in Section~\ref{core} and~\ref{lenet_weight_init}.

\begin{table}[h]
  \caption{Iterative Trimming Result on VGG-16 \{CONV5-3, FC6\}}
  \label{vgg_one_layer}
  \centering
  \begin{tabular}{llllll}
    \toprule
    \multicolumn{2}{c}{} & \multicolumn{2}{c}{Before Fine-tuning (\%)} & \multicolumn{2}{c}{After Fine-tuning (\%)} \\
    \cmidrule{3-6}
    Network & Compression & Top-1 & Top-5 & Top-1 & Top-5 \\
    (CONV5-3, FC6) & Rate & Accuracy & Accuracy & Accuracy & Accuracy \\
    \midrule
    (512, 4096)  & 1.00 & 68.36 & 88.44 & 68.36 & 88.44 \\
    (488, 3477)  & 1.19 & 64.09 & 85.90 & 71.17 & 90.28 \\
    (451, 2937)  & 1.45 & 66.77 & 87.57 & 71.08 & 90.44 \\
    (430, 2479)  & 1.71 & 68.67 & 89.17 & 71.06 & 90.34 \\
    (420, 2121)  & 1.96 & 69.53 & 89.49 & 71.05 & 90.30 \\
    (400, 1787)  & 2.28 & 68.58 & 88.92 & 70.64 & 89.97 \\
    (390, 1513)  & 2.59 & 69.29 & 89.07 & 70.44 & 89.79 \\
    \bottomrule
  \end{tabular}
\end{table}

After 6 iterations of trimming, we reduce more than half of the total number of parameters and achieve a compression rate of $2.59\times$ while the trimmed network has 2$\sim$3\% higher Top-1/Top-5 accuracy than the original VGG-16 model. The detailed performance of intermediate models are summarized in Table \ref{vgg_one_layer}. There are two interesting observations in the table. First, the initial accuracy just after trimming does not drop much from the last model even though around 500 neurons in CONV5-3 and FC6 are pruned in each iteration. This is a strong proof of redundancy in empirically designed neural networks. Also, such a small decrease in accuracy can be remedied via a fast fine-tuning instead of a week-long retraining. In our experiments, it takes less than 5K iterations to reach the original accuracy (with batch size = 256). Therefore, our trimming method allows fast optimization towards better architecture. Secondly, the trimmed networks surprisingly surpass the original VGG-16 in accuracy with less parameters. The good initialization provided by previous model sets a promising starting point for the trimmed model. In addition, having less parameters in FC6 also reduces the chance of overfitting, which may also contribute to the increment in accuracy.

\subsubsection{Trimming Multiple Layers}
VGG-16 differs from LeNet greatly in that it has a much deeper architecture with significantly more layers, which naturally gives us more options to determine which layers to trim. After the previous experiments, we want to further investigate if trimming multiple layers simultaneously can achieve the same effectiveness.

\begin{table}[h]
  \caption{Iterative Trimming Result on VGG-16 Many Layers}
  \label{vgg_many_layers}
  \begin{tabular}{llllll}
    \toprule
    \multicolumn{2}{c}{} & \multicolumn{2}{c}{Before Retraining (\%)} & \multicolumn{2}{c}{After Retraining (\%)} \\
    \cmidrule{3-6}
    Network & Compression & Top-1 & Top-5 & Top-1 & Top-5 \\
    Trimming & Rate & Accuracy & Accuracy & Accuracy & Accuracy \\
    \midrule
    no-trim baseline   & 1.00 & 68.36 & 88.44 & 68.36 & 88.444 \\
    \midrule
    CONV5-FC6-trim-1   & 1.21 & 48.18 & 72.78 & 70.58 & 90.02 \\
    CONV5-FC6-trim-2   & 1.48 & 51.18 & 77.46 & 69.80 & 89.58 \\
    CONV5-FC6-trim-3   & 1.78 & 53.58 & 79.13 & 68.84 & 89.14 \\
    \midrule
    CONV5-FC67-trim-1  & 1.23 & 47.14 & 71.72 & 70.51 & 89.97 \\
    CONV5-FC67-trim-2  & 1.54 & 50.35 & 76.54 & 69.28 & 89.29 \\
    CONV5-FC67-trim-3  & 1.88 & 51.02 & 71.65 & 68.42 & 88.80 \\
    \midrule
    CONV45-FC67-trim-1 & 1.25 & 25.96 & 46.65 & 69.55 & 89.58 \\
    CONV45-FC67-trim-2 & 1.62 & 24.79 & 48.79 & 67.97 & 89.43 \\
    CONV45-FC67-trim-3 & 2.04 & 22.85 & 45.95 & 66.20 & 87.60 \\
    \bottomrule
  \end{tabular}
\end{table}

After trimming the CONV5-3 and FC6 layers, we continue to trim their neighboring layers. We experimented with three sets of trimming layouts: \{CONV5, FC6\}, \{CONV5, FC6, FC7\}, \{CONV4, CONV5, FC6, FC7\} (see Table~\ref{vgg_many_layers}). When more neurons are pruned, the large performance drop in the trimmed network indicates retraining is necessary. We use the same set of training hyperparameters in our experiments: \{base-lr: 0.001, gamma: 0.1, step-size: 3000\}. After retraining, the trimmed networks gradually recover from the loss of neurons and rise to an accuracy level equivalent to the reference model or slightly higher. In contrast to trimming only one layer, these models regain to their capacity rather slowly, taking more than 10K iterations to recover the accuracy. 
Empirically, we found that iteratively trimming the network starting from a few layers can achieve better performance. We also found that trimming the last convolutional layer and the fully connected layers are the most effective. As shown in Table~\ref{vgg_with_fc7}, additional trimming of FC7 layer (based on previously trimmed model (CONV5-3, FC6) = (420, 2121)), can achieve a high $2.7\times$ compression rate with improved accuracy. The underlying reason is that once we have pruned the FC6 layer, the numerous zeros contribute to the high APoZ value of neurons in the FC7 layer. For the goal to reduce network parameters, it is suffices to just trim the \{CONV5-3, FC6, FC7\} layers since around $86\%$ of all the parameters are in the \{CONV5-3, FC6, FC7\} layers.


\begin{table}[h]
  \caption{Iterative Trimming Result on VGG-16 \{CONV5-3, FC6, FC7\}}
  \label{vgg_with_fc7}
  \begin{tabular}{llllll}
    \toprule
    \multicolumn{2}{c}{} & \multicolumn{2}{c}{Before Fine-tuning (\%)} & \multicolumn{2}{c}{After Fine-tuning (\%)} \\
    \cmidrule{3-6}
    Network & Compression & Top-1 & Top-5 & Top-1 & Top-5 \\
    (CONV5-3, FC6, FC7) & Rate & Accuracy & Accuracy & Accuracy & Accuracy \\
    \midrule
    (420, 2121, 4096) baseline & 1.96 & 69.53 & 89.49 & 71.05 & 90.30 \\
    \midrule
    (400, 1787, 3580) trim-1 & 2.34 & 68.15 & 88.66 & 70.55 & 90.03 \\
    (391, 1537, 3012) trim-2 & 2.70 & 68.82 & 88.86 & 70.17 & 89.69 \\
    \midrule
    (420, 2121, 3580) trim-1 & 2.00 & 70.88 & 90.11 & 71.02 & 90.30 \\
    (420, 2121, 2482) trim-2 & 2.11 & 70.21 & 89.77 & 70.88 & 90.04 \\    
    \bottomrule
  \end{tabular}
\end{table}

\section{Discussion}

\subsection{Comparison with Connection Pruning}
Work closest to ours is the work by Han et al.~\cite{learning_connections} where they iteratively prune the network connections when the correspondent weights of the connections are close to zero. They also prune a neuron when the connections to a neuron are all pruned. Compared with their work, our work is better in two major aspects. First, although Han et al. claim that they have achieved a reduction rate of parameters by $13\times$ on VGG-16, their reduction is tailored for CPU implementation of a neural network. In a GPU implementation, the convolutional layer is implemented by first vectorizing a 2D feature map into a 1D feature vector followed by a matrix multiplication~\cite{gpu-vectorize}. Thus, if a neuron is not pruned, the number of multiplications for the convolutional layers will remain the same since the vectorization is performed in a universal manner for all neurons in the same layer. This is also the same case for fully connected layers where the number of multiplications are universal for all neurons in the same layer. Note that the computational costs to re-vectorize a 2D feature map to fit for different shape of neuron connections, or adding a conditional mask checking is a lot higher than a simple matrix multiplication with redundancy. Our method, on the contrary, removes all unneeded neurons so that they do not consume any memory and are not involved in any computation at all. As shown in Section~\ref{experiment_vgg}, the trimmed VGG-16 has more than $2\times$ less FLOPs in the first fully connected layer.

Second, pruning a neuron by first pruning all of its connections is less efficient and less effective than our APoZ measurement. This is because the number of connections is significantly larger than the number of neurons in a network, especially for fully connected layers. In our experiments, we found that most of the redundancy resides in fully connected layers, and in the connections between the last convolutional layer and the first fully connected layer. However, it is rarely the case that the weights of all connections to a neuron in these layers are close to zero. Consequently, it is difficult to prune a neuron in these layers. On the other hand, our APoZ measurement can easily identify zero activation neurons for pruning regardless the weight of connections. The mean APoZ can also be used as a guideline to evaluate the effectiveness of a network as demonstrated in our experiments. 


\subsection{Dataset Used During Trimming}

In all of our experiments, we train the network using training set and run the network on validation set to obtain APoZs for neuron pruning. This method may be controversial because the validation set should not be glimpsed before finalizing the model which may potentially lead to overfitting of the validation set. We also have the same suspicion, especially after the experiments that the trimmed model can have $2\%$ higher top-5 accuracy than that of the original VGG-16 on the validation set. Therefore, we consult two more experiments to explore the potential issue.

In the first experiment, we randomly sampled a subset from training set with equal number of images (50K) as validation set. Then we used the same criteria to select weak neurons for pruning. The weak neurons selected using the sampled training set have more than $95\%$ overlap ratio with the exact neurons selected using the validation set. This shows that neurons have consistent activation performance on training and validation sets. In another word, the trimmed networks learned from sampled training data will be similar to the trimmed networks learned from the validation set.

In addition, we also tested our model on the test set of ILSVRC2012 classification track. Using single model without dense evaluation, the original VGG-16 model with $11.56\%$ validation error has an error rate of $13.02\%$ on test set. Our trimmed network with configuration \{CONV5-3: 420, FC6: 2121, FC7: 2482, Compression Rate: 2.00, Validation Error: $9.7\%$\} achieved $10.02\%$ error rate on test set. Note that the test set and validation set are non-overlapping in this ILSVRC2012 classification task. Telling from the data, after the network trimming, not only the overall accuracy is increased, but the gap between validation error and test error is also shrunk, indicating that the trimmed network has less overfitting.

The two extra experiments dismiss our concern on overfitting. They also suggest that the validation set can be used for analyzing APoZs.
 
\section{Conclusion}
We have presented Network Trimming to prune redundant neurons based on the statistics of neurons' activations. With our method, one network architecture can be deployed to handle different tasks on different datasets and the algorithm can tailor the network accordingly by determining how many neurons to use for each layer without the need of intensive computational power as well as human labor. Our method can iteratively remove low activation neurons that provide little power to the final results without damaging performance of the model. We experimented our algorithm on LeNet and VGG-16 achieving the same accuracy with 2$\sim$3$\times$ less parameters. In VGG-16, the trimmed models can even surpass the original one, which could be caused by the reduced optimization difficulty. Lying in the middle of high level network redesign and low level weight pruning, neuron pruning can be applied to any mature architecture together with weight pruning to sharply reduce the complexity of network.

\bibliographystyle{splncs}
\bibliography{egbib}

\end{document}